# Development of an automated, reliable, and clinically meaningful artificial intelligence (AI) tool for diagnosing cardiac disease from conventional cardiovascular magnetic resonance (CMR) images

Sina Amirrajab PhD[1,2], Volker Vehof MD[2], Michael Bietenbeck PhD[2], Nuriye Akyol MD[2], Redouane Bouras MD[2], Khuraman Isgandarova MD[2], Alexandru Zlibut MD[2], Philipp Stalling MD[2] and Ali Yilmaz MD[2]

(1) The D-Lab, Department of Precision Medicine, GROW - Research Institute for Oncology and Reproduction, Maastricht University, Maastricht, the Netherlands.
(2) Division of Cardiovascular Imaging, Department of Cardiology I, University Hospital Münster, Von-Esmarch-Str. 48, 48149 Münster, Germany

**ABSTRACT**

**Aims**

Cardiovascular magnetic resonance (CMR) imaging enables non-invasive assessment of myocardial structure, function, and pathology, but requires substantial experience in interpretation of CMR images that could be supported by artificial intelligence (AI)-based models. However, use of AI models for enhanced CMR reading is limited by labor-intensive data curation, suboptimal model performance, and unclear implementation pathways.

**Methods and results**

We developed an automated data curation pipeline for CMR-based cardiovascular disease (CVD) diagnosis, integrating open-source locally-run large language models (LLMs) to extract diagnostic labels from narrative CMR reports and preprocessing multimodal imaging data, including cine and late-gadolinium-enhancement (LGE) CMR sequences. Three vision foundation models (DINO, VST, UMedPT) were fine-tuned across these modalities in a two-stage approach. The dataset comprised hypertrophic cardiomyopathy (HCM), dilated cardiomyopathy (DCM), ischemic cardiomyopathy (ICM), cardiac amyloidosis (CA), and normal controls (NOR).

A total of 988 curated cases were randomly divided into 742 for training and 246 for validation. Fine-tuned AI-models achieved high discriminative diagnostic performance on an independent test set comprising 1067 patients , with individual AUC-ROC values of up to 0.937 for the correct diagnosis of HCM and 0.945 for cardiac amyloidosis. Ensemble strategies combining multiple models and modalities further improved AI-based diagnostic accuracy and robustness, achieving the highest overall diagnostic performance for HCM (AUC=0.959, CI [0.936-0.978]), CA (AUC=0.966, CI [0.939-

0.986]), NOR (AUC=0.872, CI [0.852-0.894]), DCM (AUC=0.848, CI [0.808-0.885]) and ICM (AUC=0.840, CI [0.809-0.868]).

**Conclusions**

This proof-of-concept study demonstrates that foundation model–based AI methods, when combined with automated report-based curation and multimodal CMR integration, can provide accurate, interpretable, and equitable AI-assisted cardiac diagnosis – and thereby support clinicians in correct interpretation of CMR findings.

**Abbreviations**

CMR, cardiovascular magnetic resonance; AI, artificial intelligence; CVD, cardiovascular disease; LLMs, large language models; LGE, late-gadolinium-enhancement; HCM, hypertrophic cardiomyopathy; DCM, dilated cardiomyopathy; ICM, ischemic cardiomyopathy; CA, cardiac amyloidosis; NOR, normal controls; AUC-ROC, area under the receiver operating characteristic curve; RQ, research question; 4CH, four-chamber; SA, short-axis; MLP, multilayer perceptron;

## INTRODUCTION

Cardiovascular diseases (CVDs), including heart disease and stroke, remain the leading cause of death worldwide, claiming 941,652 lives in 2022 in the United States [1]. Similarly, CVD causes over 3 million deaths every year in Europe and accounts for about 11 % of European Union healthcare expenses, placing a substantial and persistent burden on the healthcare system [2]. Cardiovascular magnetic resonance (CMR) plays a vital role in CVD by providing highly accurate, non-invasive assessment of myocardial structure, function, and tissue characterization, enabling precise diagnosis and risk stratification [3].

Despite rapid progress in artificial intelligence (AI), its clinical adoption in cardiology remains limited. Key challenges include labor-intensive data curation, suboptimal accuracy of traditional deep learning algorithms, and the lack of well-defined clinical implementation pathways.

*Foundation models*, large-scale neural networks, typically based on transformer architectures and trained on millions of data samples, represent a paradigm shift in AI development [4][5]. These models can be efficiently adapted (fine-tuned) for downstream tasks using relatively small domain-specific datasets. Examples include large language models (LLMs) for text processing [6], [7], [8] and vision foundation models such as SAM [9], MedSAM [10], and the DINO family [11], [12] for image analysis. Applications of foundation models in healthcare, particularly in CMR imaging and diagnosis, are beginning to emerge [13], [14], [15][16].

Recent studies have explored foundation models for various CMR analysis tasks. Jacob et al. [13] trained a foundation model using images from three sources comprising 27,524 subjects and reported performance on sequence and cine-view classification, segmentation, landmark localization, and disease detection from LGE images. Achieving a disease-classification accuracy of only 70%, below the state-of-the-art, their approach falls significantly short for complex CVD prediction from imaging. Moreover, zero-shot adaptation demonstrated limited effectiveness, necessitating task-specific fine-tuning with labeled data. As discussed in [14], current foundation models are not yet sufficiently mature for clinical deployment, as their performance gains remain incremental. In parallel, alternative paradigms such as reconstruction-based foundation models (e.g., the CineMA framework [16]) further illustrate the diversity of approaches being explored in this field.

Hence, a substantial gap persists in developing automated, reliable, and clinically interpretable AI tools for diagnosing CVDs from CMR imaging. In this study, we fine-tune and evaluate foundation model–based AI methods for robust and accurate cardiac disease classification from CMR images, aiming to augment clinical decision-making and improve efficiency of CMR interpretation. We address the following research questions (RQs):

RQ1: How effectively can open-source LLMs for report analysis and trained models for image processing be integrated to automate curation of CMR images for AI training in clinical practice?

RQ2: To what extent can vision foundation models be fine-tuned to achieve accurate and robust CVD classification from CMR images?

RQ3: Does an ensemble of multiple foundation models improve diagnostic accuracy and robustness in automated CMR-based CVD detection?

RQ4: Can combining multiview and multimodal CMR sequences (e.g., SA, 4CH, and LGE) enhance model performance and diagnostic confidence compared with single-view analysis?

## METHODS

### Study design and dataset

The study was conducted using anonymized, retrospective, clinical imaging data within the research landscape of the University Hospital Münster. Written broad informed consent was obtained from all participants. Hence, a specific study approval by our local Institutional Review Board (IRB) was not required. The study complies with the Declaration of Helsinki (as revised in 1975 and 2013), ICH Good Clinical Practice (GCP) guidelines, the EU Clinical Trials Regulation (EU No. 536/2014), applicable German laws and regulations (including the Medicinal Products Act), and EU/German data-protection requirements (GDPR).

The dataset comprised consecutive clinical CMR examinations acquired between 2018 and 2021 for the development cohort (training and validation) and examinations from 2022 for the independent test set. Diagnostic labels were extracted from the radiology reports using our automated LLM–based approach applied to an initial pool of approximately 8,000 examinations. For the present study, we restricted the analysis to normal controls and the four most frequent diagnostic categories to ensure adequate sample sizes per class. Cases were included based on the availability of both a) complete imaging data (cine short-axis, cine 4-chamber and LGE) and b) corresponding clinical reports and cases with mixed findings, uncertain diagnoses, or rare conditions outside the predefined categories were excluded from the image analysis.

The dataset is single-vendor, acquired on 1.5T Philips scanners (Ingenia or Ambition X models). Cine-imaging was performed using a steady-state free precession (SSFP) sequence in three long-axis slices (four-, three- and two-chamber) and a stack of short axis slices completely covering the left ventricle. LGE imaging was performed using a T1-weighted inversion recovery gradient-echo sequence 10-15 min after intravenous contrast administration (0.10 mmol/kg Gadovist®) in the same imaging planes as the cine-images.

**Automated curation pipeline**

High-quality labeled datasets are critical for successful AI development. Manual curation of CMR images is time-consuming and resource-intensive. We developed an automated data curation pipeline that uses open-source locally-run LLMs to extract diagnostic information from narrative CMR reports [17]. Multiple LLMs were used for CMR report text analysis, and their outputs were aggregated to assign diagnostic labels for patients with HCM (hypertrophic cardiomyopathy), DCM (dilated cardiomyopathy), CA (cardiac amyloidosis), ICM (ischemic cardiomyopathy), and NOR (individuals with normal cardiac findings). More precisely, we first extract the narrative section of each CMR report, which describes the primary findings and observations derived from the images. As this section was written in German, it was translated into English using Llama 3.3. The translated text was then analyzed with several locally deployed LLMs via the Ollama framework, an open-source platform that enables on-device LLM inference and thereby prevents the transfer of sensitive patient data to external servers.

We employ the three top-performing LLMs identified in our previous study [13], Gemma2-27B, DeepSeek-R1-32B, and Qwen2-5-32B, to independently analyze each report and generate diagnostic interpretations. Specifically, the LLMs independently analyzed each report and assigned a single diagnostic label from a predefined set of 38 cardiac conditions. A consensus label was then derived using a majority rule: if at least two of the three models agreed on the same diagnosis, that label was assigned to the case. Cases without agreement were not included in the final labelled dataset.

From LLM-identified cases, we selected the diagnostic categories included in this study based on having more than 100 cases per class. For the development dataset (training and validation), no manual review of the consensus labels was performed. In contrast, for the independent test set, the selected cases were reviewed by a cardiologist, and the final diagnostic labels were checked and corrected accordingly.

The normal controls (NOR) are examinations with normal left and right ventricular function and no evidence of structural or tissue abnormalities on CMR. The indications for CMR in these subjects were mostly suspected coronary artery disease (CAD) or peri-/myocarditis or cardiomyopathy work-up in patients with arrhythmias, but with ultimately normal CMR findings.

To reduce non-cardiac background and focus the analysis on the heart, corresponding short-axis (SA) late-gadolinium-enhancement (LGE), four-chamber (4CH)-cine, and SA-cine images were cropped using a previously validated bounding-box detection model [18], which avoids full segmentation while providing robust localization using a nnU-Net model [19] trained on publicly available data [18]

The overall preprocessing pipeline is illustrated in **Figure 1A.** A total of 988 curated cases were randomly divided into 742 for training and 246 for validation, which was used during model development for hyperparameter tuning and early stopping. To assess model performance, we curated an independent testing dataset comprising 1,067 patients with cardiologist-confirmed diagnoses (at least 5 years of CMR experience). A semi-automated pipeline ensured diagnostic accuracy and reproducibility of the testing data for model evaluation. Data splitting was performed at the patient level with no overlap across cohorts; follow-up examinations were excluded from the test set, and deduplication based on anonymized identifiers and DICOM metadata was applied to mitigate potential data leakage. Patient characteristics is summarized in

Table 1.

**Foundation model fine-tuning**

Three vision foundation models, DINO [11], VST [15], and UMedPT [20], were fine-tuned across three modalities, resulting in nine specialized models for cardiac disease classification **(Figure 1B)**. Fine-tuning was conducted in two stages. In the first stage, the foundation model parameters were frozen, and only the grouper module and multilayer perceptron (MLP) classification head were trained using a learning rate of $1 \times 10^{-4}$. Early stopping with a patience of 50 epochs and a cross-entropy loss function were applied to optimize the classifier parameters.

In the second stage, all model parameters were unfrozen, and training continued with a reduced learning rate of $1 \times 10^{-5}$under the same optimization scheme. The grouper module aggregates slice-level features using mean pooling across all slices to produce a single case-level representation for classification for the DINO and UMedPT models, which are 2D foundation models. For VST, a video-based Swin Transformer, the input slices are instead treated as temporal frames, and thus no grouper module is used. Empirically, this two-stage fine-tuning strategy yielded superior performance compared to end-to-end fine-tuning.

**Implementation details**

Anonymized CMR datasets were first converted from DICOM to NIfTI format for preprocessing using the TorchIO framework [21]. Images were reoriented to a standard anatomical orientation and resampled to achieve an uniform in-plane spatial resolution of 1.0 × 1.0 $mm^2$ for SA and LGE, and 1.25 × 1.25 $mm^2$ for the 4CH view. Subsequently, all images were center-cropped to a 256 × 256 matrix size, while preserving the original through-plane resolution.

A pre-trained nnU-Net backbone was employed to automatically detect bounding boxes around the heart. The resulting cropped volumes were then upsampled to 224 × 224 pixels to match the input requirements of transformer-based foundation models. For LGE, which were magnitude reconstructed images acquired at the end-diastolic phase, the volumes were cropped or padded to a fixed depth of 10 slices. From the SA stack, the center slice along with two adjacent slices above and below were selected, combined with 10 equally spaced cardiac phases sampled from the 30 available temporal frames. For the 4CH view, which contains a single slice, all 30 temporal frames were retained. Image intensities were normalized to the [0, 1] range using min–max normalization.

Data augmentation during training followed the MONAI framework [22], including random horizontal flips (p = 0.2), axial flips (p = 0.2), 90° rotations (p = 0.2), intensity scaling (p = 0.2, range = [0.8, 1.2]), and contrast adjustment (p = 0.2, range = [0.8, 1.2]). The classification head comprised a multilayer perceptron (MLP) with layer dimensions [768, 64, 5], using LeakyReLU activation functions and batch normalization after each hidden layer. Optimization was performed using the AdamW optimizer with a learning rate of $1 \times 10^{-4}$, weight decay of $1 \times 10^{-4}$, and a cosine learning rate schedule over 100 epochs, with early stopping after 50 epochs of non-improvement. All experiments were conducted on a single NVIDIA A40 GPU equipped with 48 GB of memory.

**Evaluation**

To assess the discriminative performance of the fine-tuned models, we computed the area under the receiver operating characteristic curve (AUC-ROC) for each diagnostic class. This metric quantifies the models' ability to distinguish between positive and negative cases across varying decision thresholds. In addition, we report precision, recall, F1-score, and balanced accuracy with 95% bootstrap confidence intervals for each individual model-modality combination and ensemble models. Macro-averaged performance metrics were calculated by computing each metric independently for every class using a one-vs-rest approach and then taking the unweighted mean across all classes. This approach assigns equal importance to each disease category (Normal, HCM, DCM, ICM, and CA), regardless of class size, thereby providing a balanced evaluation in the presence of class imbalance. Finally, we provide a confusion matrix for the total ensemble model to illustrate patterns of misclassification and identify disease categories that are most commonly confused.

Beyond predictive accuracy, we conducted a comprehensive evaluation following the FUTURE-AI principles [23], which define essential criteria for developing trustworthy and clinically applicable artificial intelligence systems in healthcare. Specifically, our evaluation emphasizes the fairness and explainability aspects of the foundation models.

Fairness was assessed on the independent test set by stratifying model performance according to sex and predefined age groups. For each ensemble model, AUC-ROC was calculated separately within each demographic subgroup. The sex-related fairness gap was defined as the absolute difference in AUC-ROC between male and female patients, whereas the age-related fairness gap was defined as the difference between the maximum and minimum AUC-ROC across age groups. Subgroup sample sizes were reported to aid interpretation of these comparisons. Because demographic subgroups may differ in sample size and diagnostic composition, particularly across age groups, the fairness analysis was considered exploratory and descriptive.

Explainability was assessed using gradient-based class activation mapping (Grad-CAM) [24], enabling visualization of image regions contributing most strongly to model predictions across different CMR modalities. Grad-CAM analyses were generated for representative correctly classified cases to qualitatively assess whether model attention aligned with clinically plausible disease-related imaging features. These visual explanations provide illustrative evidence supporting interpretability of the model predictions.

Together, these evaluations provide a holistic assessment of model performance and trustworthiness, addressing both diagnostic efficacy and ethical accountability in alignment with the FUTURE-AI framework.

## RESULTS

### Model performance on independent test dataset

We evaluated the performance of the models on the independent test set with confirmed diagnoses cases. The diagnostic performance of the consensus LLM-based curation approach is reported in **Supplementary Table 1**. For the fine-tuned foundation models, an overview of the nine ROC curves is presented in **Figure 2**. The best single-modality performances were achieved for the following categories: HCM (DINO–LGE, AUC = 0.937), CA (VST–SA, AUC = 0.945), NOR (UMedPT–LGE, AUC = 0.858), DCM (VST–4CH, AUC = 0.835) and ICM (DINO–LGE, AUC = 0.824).

To further enhance diagnostic accuracy, we applied an ensemble strategy by averaging predicted probabilities across models and modalities. **Figure 3** illustrates the results: the first row shows the model ensemble (averaging probabilities across modalities), the second row shows the modality ensemble (averaging across different models), and the final row presents the full ensemble that combines all models (DINO, VST, UMedPT) and all modalities (SA, 4CH, LGE), achieving the highest overall diagnostic performance with 95% confidence interval (CI) for NOR (AUC = 0. 0.872, CI [0.852, 0.894]), HCM (AUC = 0.959, CI [0.936, 0.978]), CA (AUC = 0.966, CI [0.939, 0.986]), ICM (AUC = 0.840, CI [0.809, 0.868]), DCM (AUC = 0.848, CI [0.808, 0.885]).

These results demonstrate that (1) LGE imaging provides superior diagnostic information compared to cine sequences alone, (2) ensemble methods consistently improve upon individual model performance, and (3) multi-modal integration yields the most robust classification. Comprehensive performance metrics for all ensemble models are provided in **Supplementary Table 2 and Table 3.**

At the fixed classification threshold of 0.5, the Total Ensemble showed the most balanced overall performance across diagnostic categories, with competitive macro precision, recall, F1 score, and balanced accuracy. The normalized confusion matrix demonstrated the strongest diagonal performance for HCM, normal controls, and cardiac amyloidosis, indicating higher class-wise recall for these categories. In contrast, ICM and DCM showed greater diagnostic overlap, with frequent misclassification as normal controls and additional confusion between ICM and DCM. These error patterns suggest residual phenotypic overlap between ischemic and dilated cardiomyopathy and between abnormal and normal CMR appearances in some cases (**Figure 4A**).

When ensemble strategies were compared at the same threshold, the DINO Ensemble achieved a macro F1 score comparable to the Total Ensemble, with no significant difference between the two. The Total Ensemble showed the highest macro recall and balanced accuracy, supporting its selection as the primary ensemble model. Several sequence- or architecture-specific ensembles had significantly lower macro F1 scores compared with the Total Ensemble. In particular, the SA and 4CH ensembles showed weaker threshold-based performance, suggesting that combining information across architectures and imaging sequences improved classification robustness (**Figure 4B**).

To provide qualitative insights into the ensemble model's performance, we selected representative cases from each diagnostic category. **Figure** demonstrates correctly classified examples across all five diagnostic categories, showcasing the model's ability to identify characteristic imaging patterns in multi-view CMR sequences. The total ensemble model successfully distinguished between different cardiomyopathy phenotypes in these cases. However, despite the high overall accuracy, the model exhibited some classification errors in certain challenging cases due to borderline findings (e.g. borderline size/dilation of the LV making the differentiation of DCM vs NOR challenging). **Figure 6** presents examples of misclassified patients where the ensemble prediction differed from the ground truth diagnosis.

**Fairness analysis**

The fairness analysis, summarized in **Figure 7**, evaluated ensemble model performance across sex and age subgroups. Overall, the Total Ensemble achieved the highest macro-average AUC-ROC, followed closely by the DINO, VST, and LGE ensembles. Sex-stratified analysis showed only small differences between male and female patients across all ensemble models, with female patients showing slightly higher AUC-ROC values for most ensembles. In contrast, age-stratified performance showed greater variability. The age heatmap demonstrated generally stable discrimination across most age categories, although lower AUC-ROC values were observed in some age groups for selected ensembles, particularly the 4CH ensemble.

Fairness gaps were calculated as the difference between the highest and lowest subgroup AUC-ROC within each demographic variable. Sex-related gaps were consistently small across ensembles, whereas age-related gaps were larger and varied by ensemble type. These findings suggest that model performance was relatively stable across sex subgroups but showed greater variability across age groups. Given the differences in sample size and diagnostic distribution across demographic strata, particularly the age distribution of specific diseases such as cardiac amyloidosis, these results should be interpreted as exploratory subgroup analyses rather than definitive evidence of model fairness.

**Explainability analysis**

**Figure 8** shows representative Grad-CAM overlays for the three foundation models, DINO, UmedPT, and VST, applied to LGE images across the five diagnostic categories. Across models, the activation maps were predominantly localized to the cardiac region, particularly the left ventricular myocardium and areas corresponding to visually relevant myocardial tissue. This suggests that model predictions were mainly driven by cardiac image features rather than extracardiac background structures. The spatial distribution of attention was broadly consistent across disease categories, although the extent and sharpness of activation varied between foundation models. These differences likely reflect the use of different backbone architectures and Grad-CAM target layers, which influence the spatial resolution and localization pattern of the resulting heatmaps. Overall, the explainability analysis supports that the models relied on anatomically plausible myocardial regions for classification from LGE images.

## DISCUSSION

This study demonstrates the feasibility and clinical potential of combining automated data curation with fine-tuned vision foundation models for comprehensive CMR-based cardiac disease classification. By integrating locally deployed LLMs for report interpretation with a standardized multimodal imaging pipeline, we addressed a major bottleneck in clinical AI development: the need for scalable, reliable, and privacy-preserving data labelling. The strong performance of the fine-tuned DINO, VST, and UMedPT models, together with the improvements achieved through ensemble strategies, highlights the adaptability of foundation models to complex cardiac phenotyping tasks. Importantly, the incorporation of SA cine, 4CH cine, and LGE sequences demonstrated that multimodal integration provides complementary anatomical and pathological information essential for robust disease discrimination. These findings collectively illustrate how foundation model–based pipelines can bridge the gap between algorithmic innovation and practical clinical deployment, offering a path toward more accurate, efficient, and interpretable AI-assisted cardiac diagnostics.

Building on these findings, our work systematically addresses the four research questions that guided this study:

RQ1: This study introduces an automated data curation pipeline designed for clinical environments, leveraging open-source locally-run LLMs to extract diagnostic information from narrative CMR reports and a standardized pre-processing framework to harmonize imaging data. Integrating LLM-based report parsing with pre-trained models enables automated, scalable, and accurate curation of clinical CMR datasets, substantially reducing the manual labeling burden.

RQ2: Fine-tuned vision foundation models, DINO, VST, and UMedPT, demonstrated high diagnostic accuracy across multiple cardiac disease categories, confirming the adaptability and robustness of foundation models for specialized medical imaging tasks.

RQ3: Ensemble of multiple foundation models consistently improved diagnostic robustness and accuracy by mitigating model-specific bias and variance, highlighting the benefit of integrating complementary model architectures.

RQ4: The combination of SA, 4CH views, and LGE enhanced disease discrimination and interpretability, confirming that multi-view multi-modality CMR integration captures complementary anatomical and pathological information that strengthens diagnostic performance.

The primary focus of this study was the clinical implementation of a fully automated system for data curation, the development of a pipeline for fine-tuning foundation models, and the integration of multiple modalities for CVD classification from CMR images.

A direct comparison with human expert performance was not performed and would require a prospective, multi-reader study under standardized conditions, which lies beyond the scope of the present work. In addition, comparisons across readers with different levels of experience were not performed, and therefore the potential benefit of the model in low-volume or less-experienced settings cannot be assessed based on the current data. Moreover, several of the diagnostic categories considered in this study, such as hypertrophic and dilated cardiomyopathy, are formally defined by quantitative volumetric criteria in current guidelines, which were not explicitly incorporated as structured inputs in our approach. Instead, the models implicitly capture morphological features such as ventricular dilation or hypertrophy directly from imaging data, which may explain the strong performance observed despite the absence of explicit volumetric measurements. Nevertheless, this represents an important limitation, as the model does not adhere to guideline-based thresholds and should therefore be considered as a supportive tool rather than a replacement for established diagnostic workflows.

**Clinical relevance**

Obviously, CMR is an indispensable diagnostic tool for the work-up of different cardiac diseases. However, profound experience and expertise in a) performing CMR studies and b) interpreting CMR images is required in order to achieve clinically meaningful CMR reports. Unfortunately, such a sufficient experience in reading CMR images and interpreting pathological findings is not available in some centers due to different reasons (e.g. missing specific and sufficient training in CMR, low CMR study volume, suboptimal CMR protocols, etc.). Hence, an accurate, interpretable, and equitable AI-assisted CMR analysis tool – supporting clinicians in correct interpretation of CMR findings – is highly warranted. The present study results offer a proof-of-concept approach and a preliminary solution for this important clinical demand.

In the present first study from our group, we developed an automated data curation pipeline for CMR-based CVD diagnosis, integrating open-source locally-run LLMs to extract diagnostic labels from narrative CMR reports and preprocessing multimodal, however, conventional CMR data, including SA cine, 4CH cine, and LGE sequences. This approach already enabled the correct diagnosis of common CVD such as HCM, CA, DCM and ICM with a very high diagnostic accuracy – probably outperforming resident radiologists/cardiologists with limited experience in reading CMR studies. Hence, our pipeline may be easily transformed to other CMR centers and be used for supporting clinicians in correct interpretation of CMR findings.

As a next step, we intend to increase the number of cardiac disease groups in order to cover the majority of CVD that can be detected by CMR. To achieve this next goal, increasing the size of the dataset will not be sufficient. Future approaches will include integration of additional clinical data, incorporating other CMR views (e.g. three chamber-view, 3CH), functional parameters (ventricular volumes, ejection fraction, cardiac output), tissue characterization metrics (T1, T2, ECV mapping), and demographic or clinical variables (age, sex, comorbidities) to improve model precision, especially in underperforming disease categories**.** Fortunately, the present AI-pipeline can be easily adopted to these additional requirements and thereafter quickly tested and validated.

In the current study, the model is trained and evaluated strictly on the five predefined diagnostic categories (HCM, DCM, ICM, CA), and therefore all inputs are assigned to one of these classes. Future work could explore mechanisms to identify out-of-distribution cases, for example by training an additional “unknown” class or using uncertainty estimation, to improve the model’s safety and reliability in clinical deployment. Additionally, while fine-tuning on the most prevalent diagnostic categories enables robust performance for common conditions, this approach may limit generalizability to rare or underrepresented diseases. Expanding the label space and evaluating performance on long-tail distributions will be an important next step to improve clinical applicability.

**Limitations**

A direct comparison of our approach with the extra-ordinary work of Wang et al. [15] was not possible due to the unavailability of their trained model weights – that were unfortunately not shared with us by these authors. In this context, we note that the AUC values reported by Wang et al. [15] are extremely high (approaching 1.0) and should be interpreted carefully. Such high values may reflect differences in cohort composition, inclusion/exclusion criteria, disease spectrum, task definition – and potential exclusion of borderline cases. However, since we do not have any detailed insight into the respective data, we are forced to believe that Wang et al. looked at a consecutive cohort including mild/borderline cases and cases beyond the primary diagnostic label space. Using the same VST architecture as in their study, we fine-tuned the model on our dataset and obtained strong performance that we consider more representative of routine clinical practice. Evaluation on external public datasets, such as ACDC [25], M&Ms 1 [26] and 2 [27], was not feasible because paired SA, 4CH, and LGE images were unavailable, and these datasets do not encompass a comprehensive set of diagnostic classes used in this study. To facilitate reproducibility and further research, we have open-sourced the code, trained weights, and inference scripts.

A key limitation of this work is the reliance on single-center data, which may restrict the generalizability of the models to multi-center or heterogeneous clinical settings. Future efforts will focus on expanding the dataset across institutions to enhance robustness and external validity.

Another limitation is the use of only three central short-axis slices, which may not capture pathological findings present in more basal or apical regions. While this design balances information content and computational efficiency, the optimal number of slices and cardiac phases for reliable diagnosis was not systematically evaluated and remains to be determined in future work.

A further limitation is that we did not directly quantify the reliability of LLM-derived labels in the training and validation cohorts through manual expert review. Although the development labels were assigned using a consensus approach across multiple LLMs, this does not eliminate the risk of systematic mislabeling, as different models may converge on the same incorrect diagnosis. While the independent test set was manually verified by a cardiologist and the performance of the LLM-based labeling approach was evaluated against cardiologist-confirmed diagnoses, the absence of a dedicated audit of the development cohort limits our ability to fully estimate potential label noise introduced by automated translation and report interpretation. Ambiguous report language, uncertain diagnoses, and mixed phenotypes may further contribute to systematic labeling errors. Future work should therefore include clinician review of representative subsets of automatically labeled cases to quantify agreement and better characterize common error patterns within the LLM-based curation pipeline.

**Conclusion**

This proof-of-concept study demonstrates that foundation model–based AI methods, when combined with automated report-based curation and multimodal CMR integration, can provide accurate, interpretable, and equitable AI-assisted cardiac diagnosis – and thereby support clinicians in correct interpretation of CMR findings. Ensemble approaches enhance robustness, supporting scalable clinical deployment, reducing manual labeling burden, and improving decision-making in cardiovascular care.

**Software code**

All training and inference code, along with the trained model weights, are publicly available on https://github.com/sinaamirrajab/CMR_CVD to support full reproducibility of the experiments.

**Declarations**

- Ethics approval and consent to participate: The study protocol complies with the Declaration of Helsinki. Written informed broad consent was obtained from the participants.
- Consent for publication: N/A.
- Availability of data and materials: The datasets used and/or analyzed during the current study are available from the corresponding author on reasonable request.

**Funding sources**

This research project is funded by a research grant from the Peter-Lancier-Stiftung (Hamburg, Germany).

**Authors' contributions**

SA designed the study, wrote major paragraphs of the manuscript, critically reviewed the final manuscript and drafted the manuscript. VV and MB participated in the CMR exams, carried out the data analysis, and wrote some paragraphs of the draft version of the manuscript. NA, RB, KI and PS participated in the CMR exams and in the analysis of the CMR data. AY designed and supervised the study, wrote major paragraphs of the manuscript and critically reviewed the final manuscript. All authors read and approved the final manuscript.

*Table 1 Patient Characteristics and Disease Distribution*

| | Train | Validation | Test |
|---|---|---|---|
| **Patients, n** | 742 | 246 | 1067 |
| **Age, mean± SD** | 46.3 ± 19.5 | 47.0 ± 19.5 | 51.7 ± 17.1 |
| **Sex: Male, n (%)** | 474 (63.9%) | 159 (64.6%) | 662 (62.0%) |
| **Sex: Female, n (%)** | 268 (36.1%) | 87 (35.4%) | 405 (38.0%) |
| **Study year range** | 2018-2022 | 2018-2022 | 2022 |
| **NOR** | 256 (34.5%) | 79 (32.1%) | 636 (59.6%) |
| **DCM** | 168 (22.6%) | 57 (23.2%) | 112 (10.5%) |
| **HCM** | 115 (15.5%) | 40 (16.3%) | 76 (7.1%) |
| **ICM** | 113 (15.2%) | 38 (15.4%) | 175 (16.4%) |
| **CA** | 90 (12.1%) | 32 (13.0%) | 68 (6.4%) |

*Abbreviations: HCM, hypertrophic cardiomyopathy; DCM, dilated cardiomyopathy; ICM, ischemic cardiomyopathy; CA, cardiac amyloidosis; NOR, normal control.*

**FIGURE LEGENDS:**

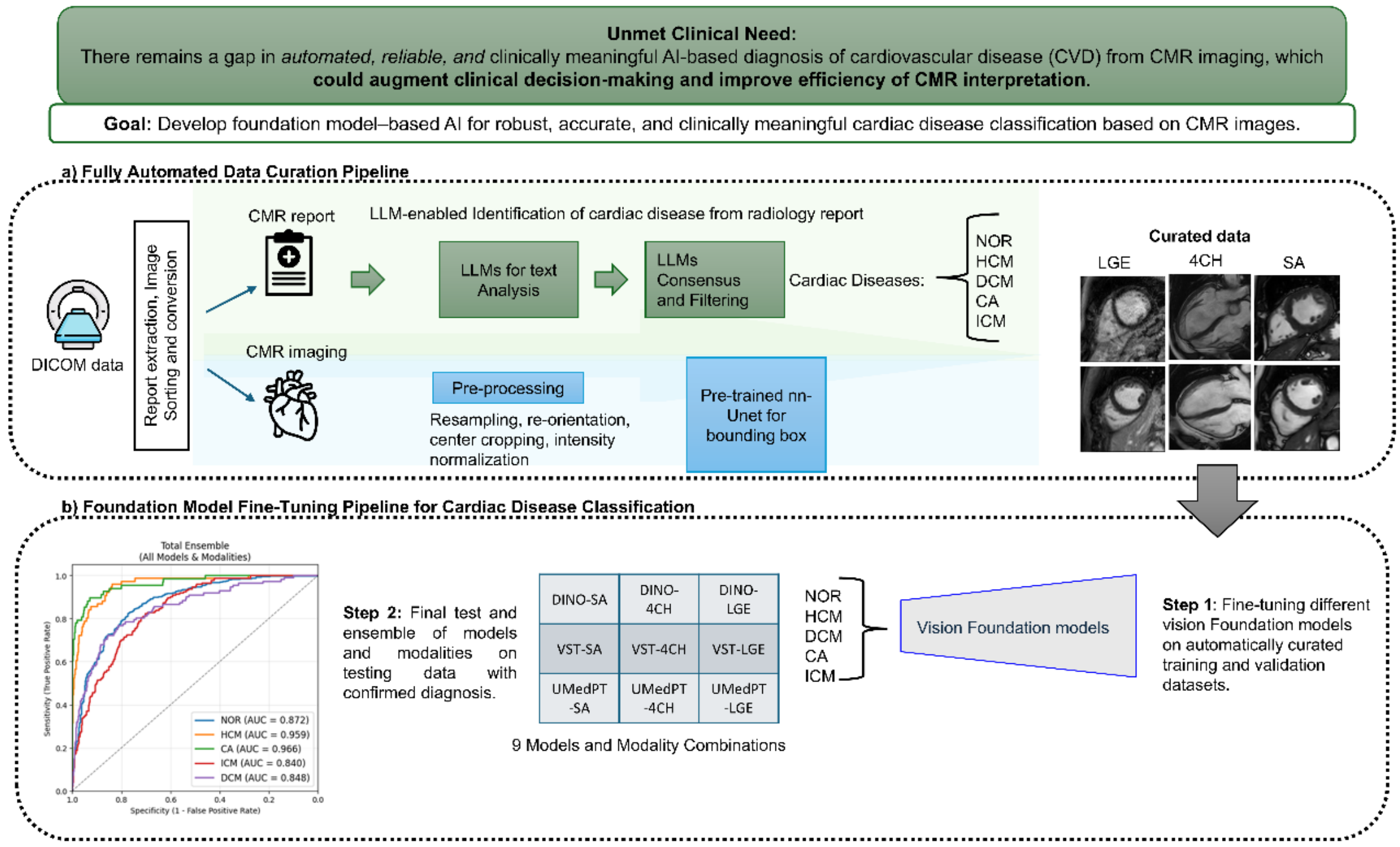

**Graphical abstract:** The overview of the fully automated data curation pipeline (a), which includes locally run open-source Large Language Models (LLMs) for extracting diagnostic information from cardiovascular magnetic resonance (CMR) imaging reports and automated processing of late-gadolinium enhancement (LGE), cine short-axis (SA), and four-chamber (4CH) images. The foundation model fine-tuning pipeline for cardiac disease classification (b) is trained and validated on the curated data and evaluated on an independent test set with confirmed diagnosis cases of hypertrophic cardiomyopathy (HCM), dilated cardiomyopathy (DCM), cardiac amyloidosis (CA), ischemic cardiomyopathy (ICM), and normal controls (NOR). Combining different foundation models and imaging modalities achieves the highest diagnostic performance based on the area under the receiver operating characteristic (AUC-ROC) metric.

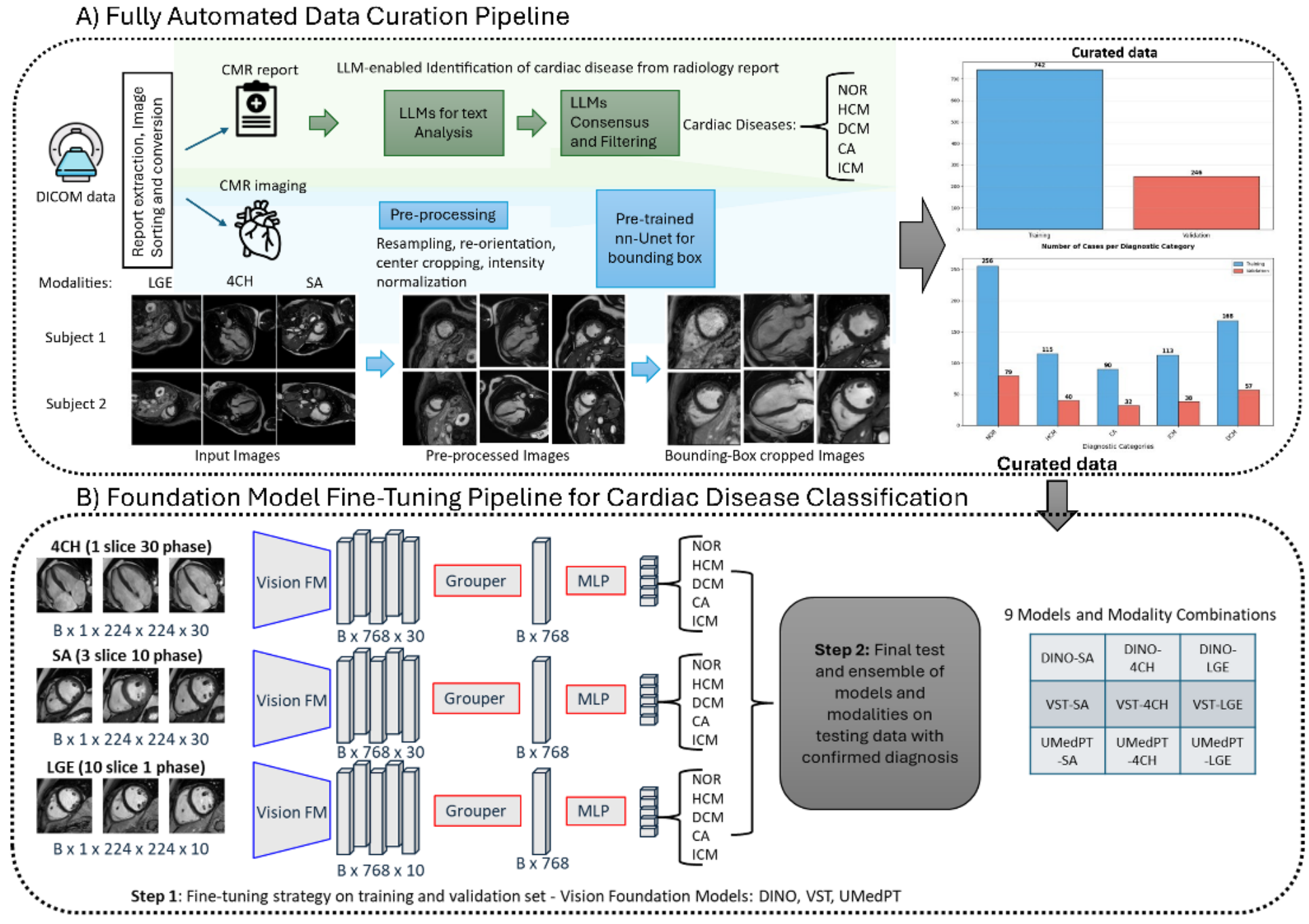

**Figure 1** The overview of the fully automated data curation pipeline (a) and foundation model fine-tuning pipeline for cardiac disease classification (b).

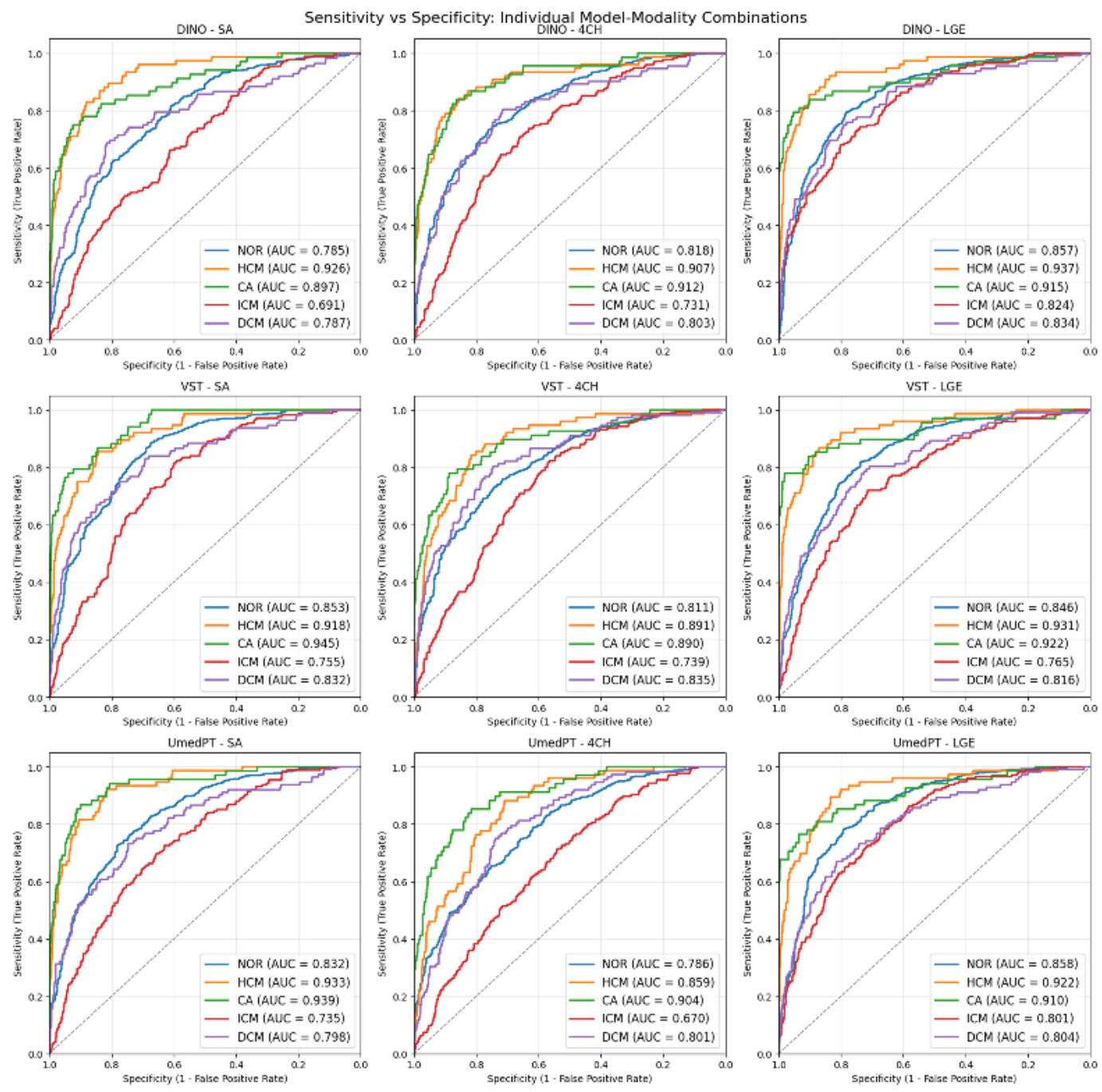

**Figure 2** ROC curves and AUC performance of different models and modalities on the curated unseen testing data with clinically confirmed diagnostic classes.

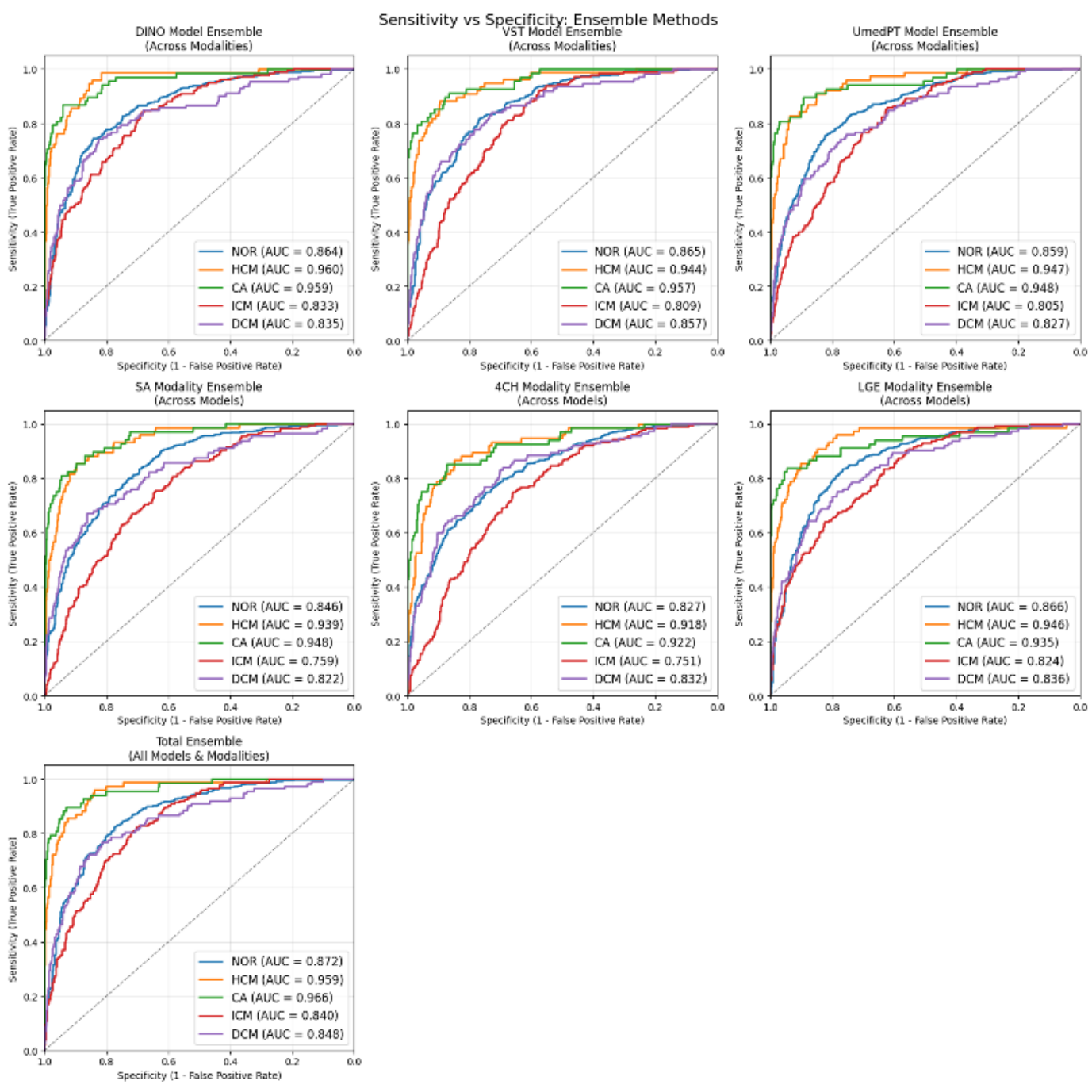

**Figure 3** Ensemble of different foundation models across all modalities (first row), ensemble of different modalities across all foundation models (second row), and ensemble of all models (DINO, VST, UmedPT) and modalities (SA, 4CH, LGE) achieves the highest diagnostic performance (third row).

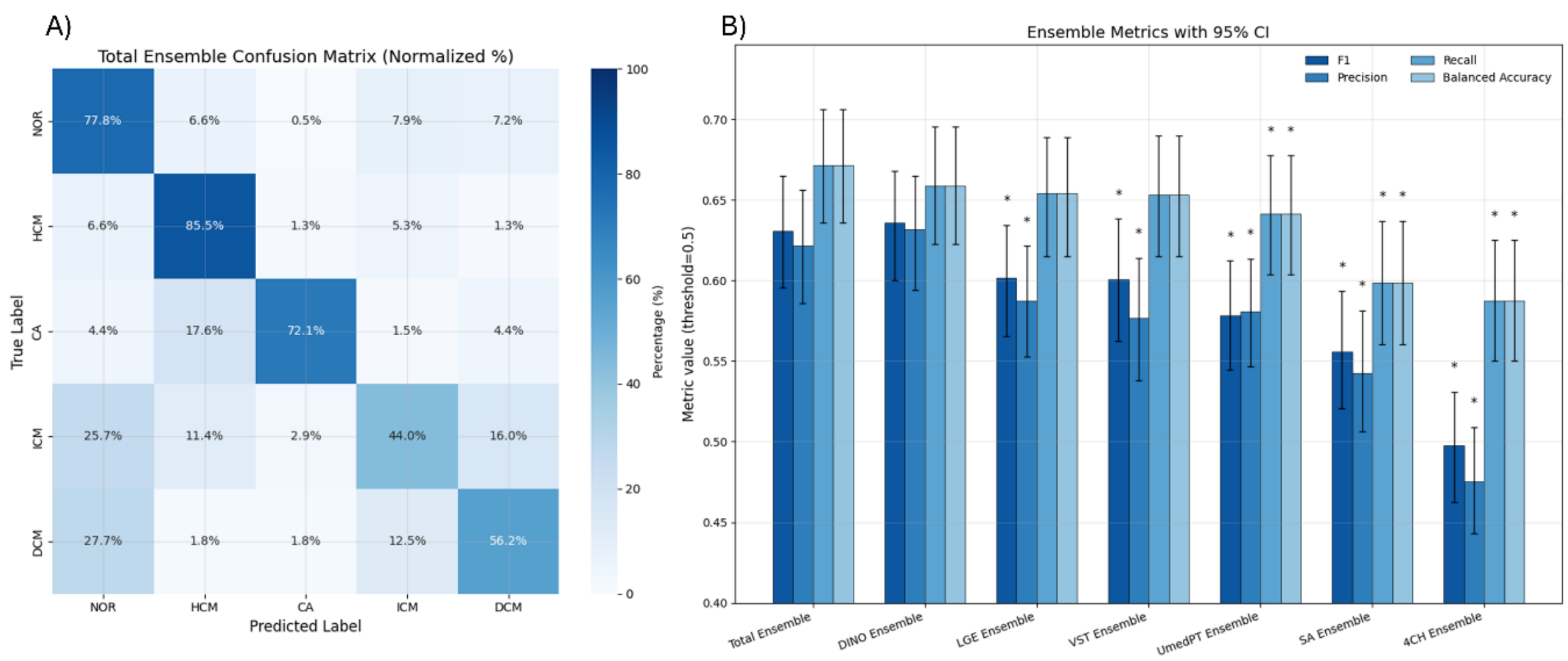

**Figure 4** Threshold-based performance of ensemble models. A) Normalized confusion matrix for the Total Ensemble at a fixed classification threshold of 0.5. Diagonal values represent per-class recall. The strongest classification performance was observed for HCM, normal controls, and cardiac amyloidosis, whereas the largest residual confusions occurred between ICM and normal controls, DCM and normal controls, and between ICM and DCM. B) Ensemble-level macro F1 score, precision, recall, and balanced accuracy with 95% confidence intervals at a fixed threshold of 0.5. Confidence intervals were estimated using 1,000 bootstrap resamples. Asterisks indicate statistically significant differences compared with the Total Ensemble.

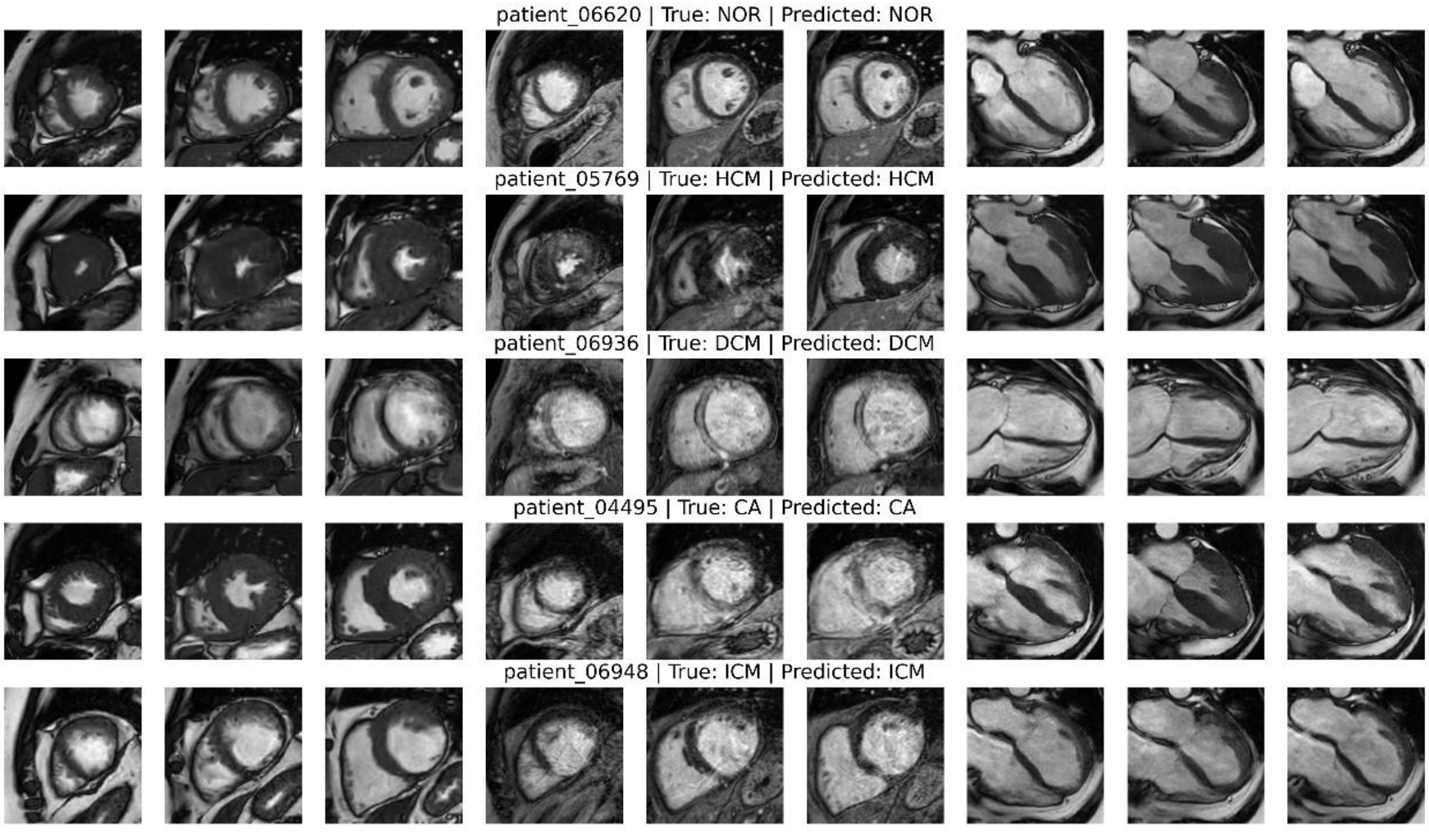

**Figure 5** Representative examples of correctly classified cardiac MRI images by the total ensemble model across five diagnostic categories. Each row shows multi-view cardiac MRI images from a single patient, with the ensemble model successfully predicting: (A) Normal (NOR), (B) Hypertrophic Cardiomyopathy (HCM), (C) Cardiac Amyloidosis (CA), (D) Ischemic Cardiomyopathy (ICM), and (E) Dilated Cardiomyopathy (DCM).

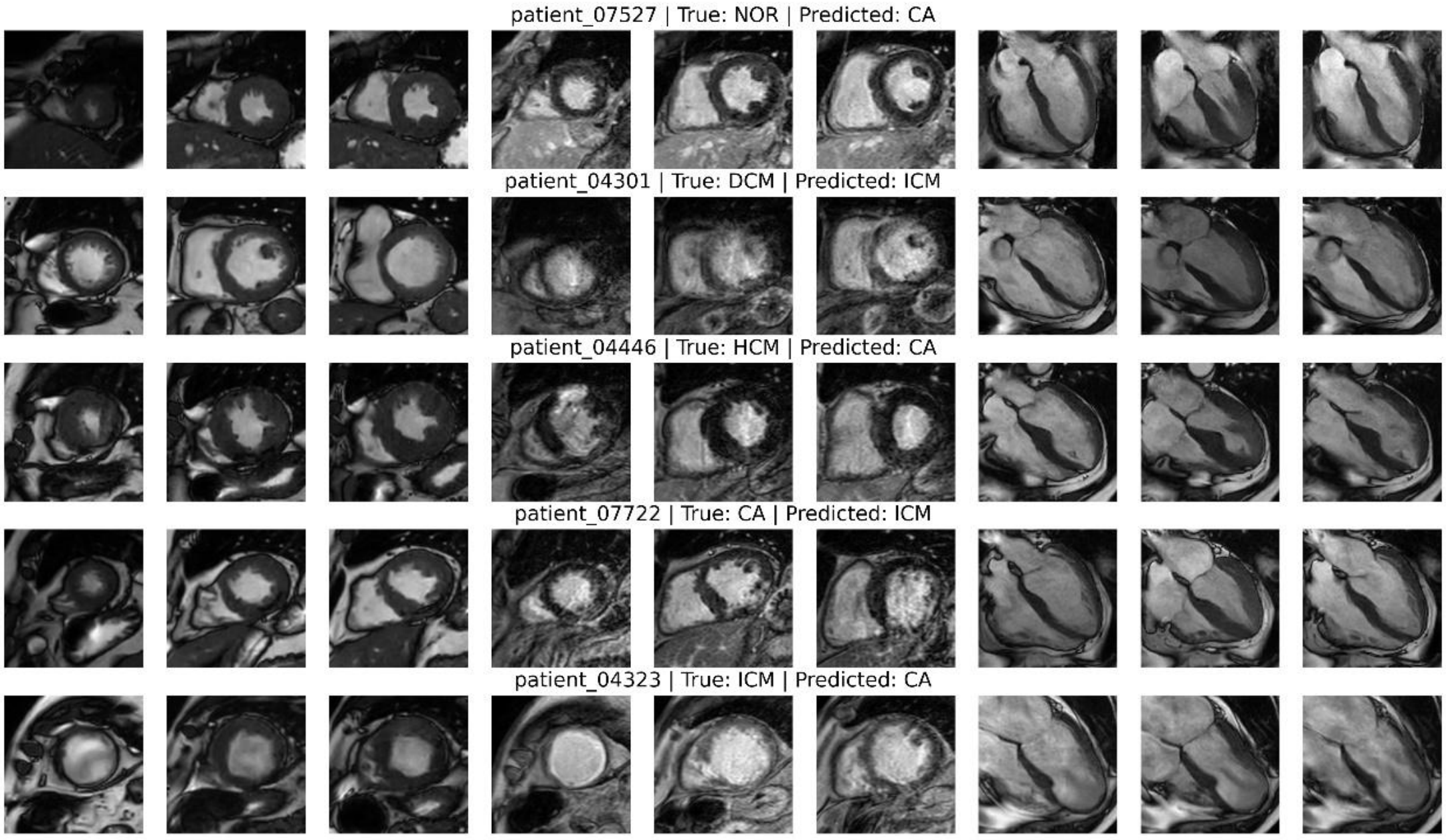

**Figure 6** Examples of misclassified cases showing true versus predicted diagnostic categories. These challenging cases illustrate the difficulty in distinguishing certain cardiomyopathy phenotypes based on imaging features alone.

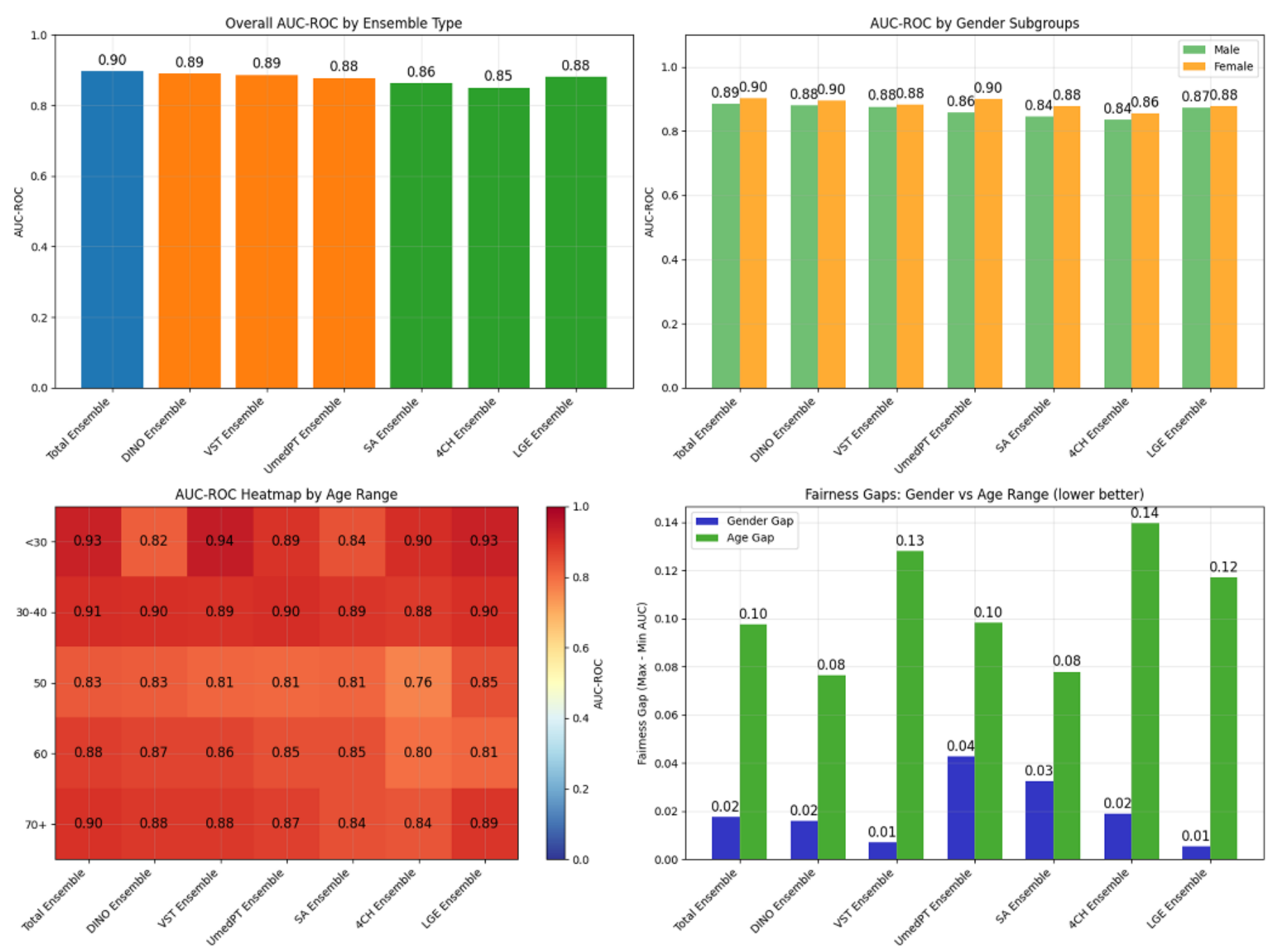

**Figure 7** Comprehensive Fairness Analysis of Ensemble Models, evaluating the performance and fairness of ensemble models across demographic subgroups. Top Left: Overall AUC-ROC scores by ensemble type. Top Right: Gender subgroup AUC-ROC scores with average sample sizes in the legend. Bottom Left: AUC-ROC heatmap by age range with sample sizes annotated. Bottom Right: Fairness gap analysis (Gender vs. Age).

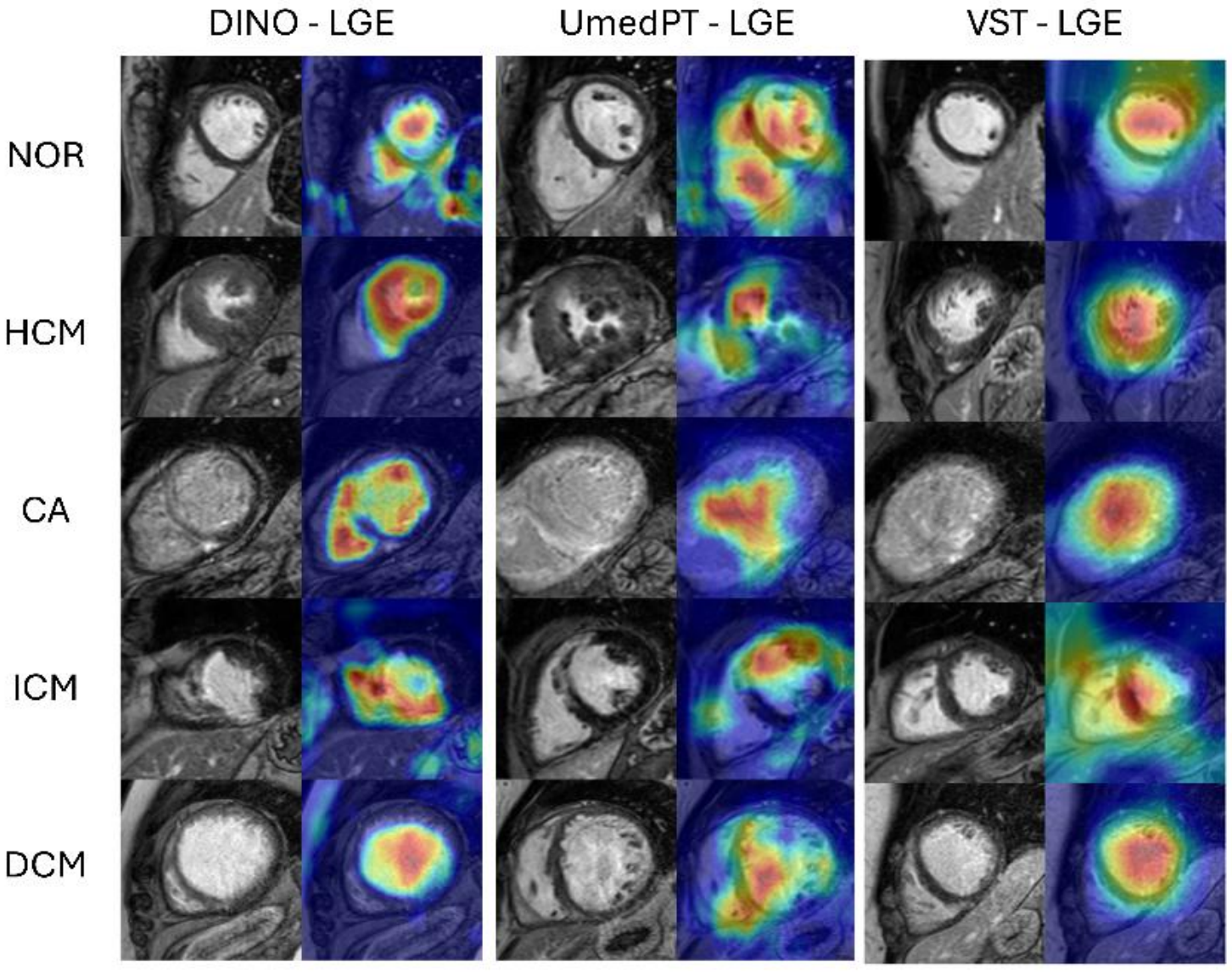

**Figure 8**. Grad-CAM explainability analysis for LGE-based classification. Representative LGE slices and corresponding Grad-CAM overlays are shown for DINO, UmedPT, and VST across normal controls, HCM, cardiac amyloidosis, ICM, and DCM. Activation maps were mainly localized to the cardiac region and left ventricular myocardium, with limited attention to extracardiac background structures. Differences in the extent and sharpness of activation reflect model-specific backbone architectures and Grad-CAM layer selection.